\documentclass[conference]{IEEEtran}
\IEEEoverridecommandlockouts
\usepackage{cite}
\usepackage{amsmath,amssymb,amsfonts}
\usepackage{algorithmic}
\usepackage{graphicx}
\usepackage{textcomp}
\usepackage{xcolor}
\usepackage{multirow}
\def\BibTeX{{\rm B\kern-.05em{\sc i\kern-.025em b}\kern-.08em
    T\kern-.1667em\lower.7ex\hbox{E}\kern-.125emX}}
\begin{document}

\title{Comprehend Medical: a Named Entity Recognition and Relationship Extraction Web Service}

\author{
     \IEEEauthorblockN{Parminder Bhatia}
     \IEEEauthorblockA{
         \textit{Amazon} \\
         Seattle, Washington, USA  \\
         parmib@amazon.com
     }
     \and
     \IEEEauthorblockN{Busra Celikkaya}
     \IEEEauthorblockA{
         \textit{Amazon} \\
         Seattle, Washington, USA  \\
         busrac@amazon.com
      }
     \and
     \IEEEauthorblockN{Mohammed Khalilia}
     \IEEEauthorblockA{
         \textit{Amazon} \\
         Seattle, Washington, USA  \\
         khallia@amazon.com
     }
     \and
     \IEEEauthorblockN{Selvan Senthivel}
     \IEEEauthorblockA{
         \textit{Amazon} \\
         Seattle, Washington, USA  \\
         ssenthiv@amazon.com
     }
 }

\maketitle

\begin{abstract}
Comprehend Medical is a stateless and Health Insurance Portability and Accountability Act (HIPAA) eligible Named Entity Recognition (NER) and Relationship Extraction (RE) service launched under Amazon Web Services (AWS) trained using state-of-the-art deep learning models. Contrary to many existing open source tools, Comprehend Medical is scalable and does not require steep learning curve, dependencies, pipeline configurations, or installations. Currently, Comprehend Medical performs NER in five medical categories: Anatomy, Medical Condition, Medications, Protected Health Information (PHI) and Treatment, Test and Procedure (TTP). Additionally, the service provides relationship extraction for the detected entities as well as contextual information such as negation and temporality in the form of traits. Comprehend Medical provides two Application Programming Interfaces (API): 1) the NERe API which returns all the extracted named entities, their traits and the relationships between them and 2) the PHId API which returns just the protected health information contained in the text. Furthermore, Comprehend Medical is accessible through AWS Console, Java and Python Software Development Kit (SDK), making it easier for non-developers and developers to use.
\end{abstract}


\section{Introduction}
Electronic Health Records (EHR) contain a wealth of patients' data ranging from diagnoses, problems, treatments, medications to imaging and clinical narratives such as discharge summaries and progress reports. Structured data are important for billing, quality and outcomes. On the other hand, narrative text is more expressive, more engaging and captures patient's story more accurately. Narrative notes may also contain information about level of concern and uncertainty to others who are reviewing the note. Studies have shown that narrative notes contain more naturalistic prose, more reliable in identifying patients with a given disease and more understandable to healthcare providers reviewing those notes \cite{Rosenbloom2011, Fox1998, Marill1999, VanGinneken1996, Cawsey1997}. Therefore, to have a clear perspective on patient condition, narrative text should be analyzed. However, manual analysis of massive number of narrative text is time consuming, labor intensive and prone to errors.

Many clinical Natural Language Processing (NLP) tools and systems were published to help us make sense of those valuable narrative text. For instance, clinical Text Analysis and Knowledge Extraction System (cTAKES) \cite{Savova2010} is an open-source NLP package based on the Unstructured Information Management Architecture (UIMA) framework \cite{FERRUCCI2004} and OpenNLP \cite{opennlp} natural language processing toolkit. cTAKES uses a dictionary look-up and each mention is mapped to a Unified Medical Language System (UMLS) concept \cite{umls}. MetaMap \cite{Aronson2010} is another open-source tool aims at mapping mentions in biomedical text to UMLS concepts using dictionary lookup. MetaMap Lite \cite{Demner-Fushman} adds negation detection based on either ConText \cite{Harkema2009} or NegEx \cite{negex}.

The Clinical Language Annotation, Modeling, and Processing (CLAMP) \cite{Soysal2018} is one of the most recent clinical NLP systems. CLAMP is motivated by the fact that existing clinical NLP systems need customization and must be tailored to one's task. For NER, CLAMP takes two approaches: machine learning approach using Conditional Random Field (CRF) \cite{lafferty2001conditional} and dictionary-based, which maps mentions to standardized ontologies. CLAMP also provides assertion and negation detection based on machine learning or rule-based NegEx.

Many of the existing NLP systems rely on ConText \cite{Harkema2009} and NegEx \cite{negex} to detect assertions such as negation. ConText extracts three contextual features for medical conditions: negation, historical or hypothetical and experienced by someone other than the patient. ConText is an extension of NegEx, which is based on regular expression.

Most of the NLP systems discussed above perform linking of mentions to UMLS. They are based on pipelined components that are configurable, rely on dictionary look-up for NER and regular expressions for assertion detection.

Recently, neural network models have been proposed to overcome some of the limitations of rule-based techniques. A feedforward and bidirectional Long Short Term Memory (BiLSTM) networks for generic negation scope detection was proposed in \cite{fancellu2016neural}. In \cite{rumeng2017hybrid} a gated recurrent units (GRUs) are used to represent the clinical relations and their context, along with an attention mechanism. Given a text annotated with relations, it classifies the presence and period of the relations. However, this approach is not end-to-end as it does not predict the relations. Additionally, these models generally require large annotated corpus to achieve good performance, but clinical data is scarce.

Kernel-based approaches are also very common, especially in the 2010 i2b2/VA task of predicting assertions. The state-of-the-art in that challenge applied support vector machines (SVM) to assertion prediction as a separate step after entity extraction \cite{de2011machine}. They train classifiers to predict assertions of each concept word, and a separate classifier to predict the assertion of the whole entity. Augmented Bag of Words Kernel (ABoW), which generates features based on NegEx rules along with bag-of-words features was proposed in \cite{shivade2015extending} and a CRF based approach for classification of cues and scope detection was proposed in \cite{cheng2017automatic}. These machine learning based approaches often suffer in generalizability.

Once named entities are extracted it is important to identify the relationships between the entities. Several end-to-end models were proposed that jointly learn named entity recognition and relationship extraction \cite{miwa2016end, zheng-etal-2017-joint, adel-schutze-2017-global}. Generally, relationship extraction models consist of an encoder followed by relationship classification unit \cite{verga-etal-2018-simultaneously, christopoulou-etal-2018-walk, su-etal-2018-global}. The encoder provides context aware vector representations for both target entities, which are then merged or concatenated before being passed to the relation classification unit, where a two layered neural network or multi-layered perceptron classifies the pair into different relation types.

Despite the existence of many clinical NLP systems, automatic information extraction from narrative clinical text has not achieved enough traction yet \cite{Wang2018}. As reported by \cite{Wang2018} there is a significant gap between clinical studies using Electornic Health Record (EHR) data and studies using clinical information extraction. Reasons for such gap can be attributed to limited expertise of NLP experts in the clinical domain, limited availability of clinical data sets due to the HIPAA privacy rules and poor portability and generalizability of clinical NLP systems. Rule-based NLP systems require handcrafted rules, while machine learning-based NLP systems require annotated datasets.

To narrow the clinical NLP adoption gap and to address some of the limitations in existing NLP systems, we present Comprehend Medical, a web service for clinical named entity recognition and relationship extraction. Our contributions are as follows: 

\begin{itemize}
\item Named entity recognition, relationship extraction and trait detection service encapsulated in one easy to use API.  
\item Web service that uses deep learning multi-task \cite{bhatia2019dynamic} approach trained on labeled training data and requires no configurations or customization.
\item Trait (negation, sign, symptom and diagnosis) detection for medical condition and negation detection for medication.
\end{itemize}

The rest of the paper is organized as follows: section \ref{sec:methods} presents the methods, section \ref{sec:exp} describes the datasets and experimental settings, section \ref{sec:results} contains the results for the NER and RE models, section \ref{sec:imp} talks about the implementation details, section \ref{sec:entities} gives overview of the supported entities, traits and relationships, section \ref{sec:use_cases} presents some of the use cases and we conclude in section \ref{sec:conclusion}.

\section{Methods}
\label{sec:methods}
In this section we briefly introduce the architectures for named entity recognition and trait detection proposed in \cite{Bhatia2019} and the relation extraction using explicit context conditioning proposed in \cite{gaurav2019}.

\subsection{Named Entity Recognition Architecture}\label{nerarch}
A sequence tagging problem such as NER can be formulated as maximizing the conditional probability distribution over tags $\mathbf{y}$ given an input sequence $\mathbf{x}$, and model parameters $\theta$. 
\begin{equation}
    P(\mathbf{y} | \mathbf{x}, \theta) = {\displaystyle \prod_{t=1}^{T} P(y_t | x_t, y_{1:t-1}, \theta)}
\end{equation}
$T$ is the length of the sequence, and $y_{1:t-1}$ are tags for the previous words.
The architecture we use as a foundation is that of \cite{lample2016neural,yang2016multi}.
The model consists of three main components: (i) character encoder, (ii) word encoder, and (iii) decoder/tagger.

\subsubsection{Encoders}
Given an input sequence $\mathbf{x} \in \mathbb{N}^T$ whose coordinates indicate the words in the input vocabulary, we first encode the character level representation for each word.
For each $x_t$ the corresponding sequence $\mathbf{c}^{(t)} \in \mathbb{R}^{L \times e_c}$ of character embeddings is fed into an encoder, where $L$ is the length of a given word and $e_c$ is the size of the character embedding.
The character encoder employs two LSTM  units which produce $\overrightarrow{h^{(t)}_{1:l}}$, and $\overleftarrow{h^{(t)}_{1:l}}$, the forward and backward hidden representations, respectively, where $l$ is the last timestep in both sequences.
We concatenate the last timestep of each of these as the final encoded representation, $h_c^{(t)} = [\overrightarrow{h^{(t)}_l} || \overleftarrow{h^{(t)}_l}]$, of $x_t$ at the character level.

The output of the character encoder is concatenated with a pre-trained word embedding, $m_t = [h_c^{(t)} || \text{emb}_{word}(x_t)]$, which is used as the input to the word level encoder.

Using learned character embeddings alongside word embeddings has shown to be useful for learning word level morphology, as well as mitigating loss of representation for out-of-vocabulary words.
Similar to the character encoder we use a BiLSTM to encode the sequence at the word level.
The word encoder does not lose resolution, meaning the output at each timestep is the concatenated output of both word LSTMs, $h_t = [\overrightarrow{h_t} || \overleftarrow{h_t}]$.

\subsubsection{Decoder and Tagger}
Finally, the concatenated output of the word encoder is used as input to the decoder, along with the label embedding of the previous timestep.  
During training we use teacher forcing \cite{Williams1989} to provide the gold standard label as part of the input.
\begin{equation}
    o_t = \text{LSTM}(o_{t-1}, [h_t || \hat{y}_{t-1}])
\end{equation}
\begin{equation}
    \hat{y}_t = \text{Softmax}(\mathbf{W}o_t + b^s)
\end{equation}
where $\mathbf{W} \in \mathbb{R}^{d \times n}$, $d$ is the number of hidden units in the decoder LSTM, and $n$ is the number of tags.
The model is trained in an end-to-end fashion using a standard cross-entropy objective.

\subsubsection{Named Entity Recognition Decoder Model} 
\begin{figure}[htbp]
\centering
    \centerline{\includegraphics[scale=0.4]{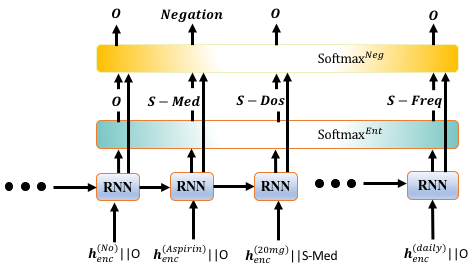}}
    \caption{Conditional softmax decoder model}
    \label{fig:conditional_softmax}
\end{figure}

Our decoder model provides more context to trait detection by adding an additional input, which is the softmax output from entity extraction. We refer to this architecture as the Conditional Softmax Decoder as shown in Fig. \ref{fig:conditional_softmax} \cite{Bhatia2019}. Thus, the model learns more about the input as well as the label distribution from entity extraction prediction. As an example, we use negation only for problem entity in the i2b2 dataset. Providing the entity prediction distribution helps the negation model to make better predictions. The negation model learns that if the prediction probability is not inclined towards the problem entity, then it should not predict negation irrespective of the word representation.
\begin{equation}
    \hat{y}^{Entity}_t, \text{SoftOut}^{Entity}_t = \text{Softmax}^{Ent}(\mathbf{W^{Ent}}o_t + b^s)
\end{equation}
\begin{equation}
    \hat{y}^{Neg}_t = \text{Softmax}^{Neg}(\mathbf{W^{Neg}}[o_t,\text{SoftOut}^{Entity}_t] + b^s)
\end{equation}
where, $\text{SoftOut}^{Entity}_t$ is the softmax output of the entity at time step $t$. 

Readers are referred to \cite{Bhatia2019} for more detailed discussion on the conditional softmax decoder model.

\subsection{Relationship Extraction Architecture}
The extracted entities are not very meaningful by themselves, specially in the healthcare domain. For instance, it is important to know if the procedure was performed bilaterally, on the left or right side. Knowing the correct location will result in more accurate and reliable billing and reimbursement. Hence, it is important to identify the relationships among those clinical entities.

\begin{figure}[htbp]
\centering
    \centerline{\includegraphics[scale=0.4]{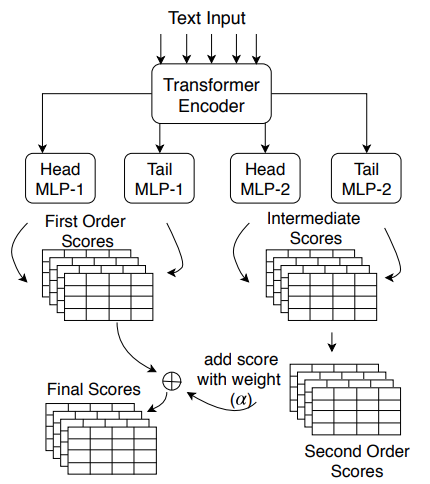}}
    \caption{Relationship extraction model}
    \label{fig:rel_model}
\end{figure}

The RE model architecture is described in \cite{gaurav2019}, but we reiterate some of the important details here.  Relationships are defined between two entities, which we refer to as head and tail entity. To extract such relationships we proposed relation extraction using explicit context conditioning, where two target entities (head and tail) can be explicitly connected via a context token also known as second order relations. Similar to Bi-affine Relation Attention Networks (BRAN) \cite{verga-etal-2018-simultaneously}, we first compute the representations for both the head, $e_{i}^{head}$, and tail, $e_{i}^{tail}$, entities, which are then passed through two multi-layer perceptron (MLP-1) to obtain first-order relation scores, $score^{(1)}(p^{head}, p^{tail})$, as shown in Fig. \ref{fig:rel_model}. We also pass $e_{i}^{head}$ and $e_{i}^{tail}$ through MLP-2 to obtain second-order relation scores, $score^{(2)}(p^{head}, p^{tail})$, where $p^{head}$ and $p^{tail}$ are the indices for the head and tail entities. The motivation for adding MLP-2 was driven by the need for representations focused on establishing relations with context tokens, as opposed to first-order relations. At the end, the final score for relation between two entities is given as a weighted sum of first and second order scores.

\section{Experiments}
\label{sec:exp}
\subsection{Dataset}
We evaluated our model on two datasets. First is the 2010 i2b2/VA challenge dataset for ``test, treatment, problem'' (TTP) entity extraction and assertion detection, herein referred to as \textit{i2b2}. Unfortunately, only part of this dataset was made public after the challenge, therefore we cannot directly compare with NegEx and ABoW results. We followed the original data split from \cite{chal2016ner} of 170 notes for training and 256 for testing. The second dataset is proprietary and consists of  4,200 de-identified clinical notes with medical conditions, herein referred to as \textit{DCN}.

The i2b2 dataset contains six predefined relations types including \textit{TrCP} (Treatment Causes Problem), \textit{TrIP} (Treatment Improves Problem), \textit{TrWP} (Treatment Worsens Problem) and one negative relation. The DCN dataset contains seven predefined relationship types such as \textit{with\_dosage}, \textit{every} and one negative relation.
A summary of the datasets is presented in Table \ref{tab:datasets}.

\begin{table}[ht!]
  \caption{Overview of the i2b2 and DCN datasets.}
  \label{tab:datasets}
  \begin{center}
  \begin{tabular}{|l|l|l|}
  \hline
  &\textbf{i2b2}&\textbf{DCN} \\
  \cline{1-3} 
    Notes & 426 & 4,200 \\
    Tokens  & 416K & 1.5M \\
    Entity Tags  & 13 & 37 \\
    Relations & 3,653 & 270,000 \\
    Relation Types & 6 & 7 \\
    \hline
\end{tabular}
\end{center}
\end{table}

\subsection{NER Model Settings}
Word, character and tag embeddings are 100, 25, and 50 dimensions, respectively. Word embeddings are initialized using GloVe, while character and tag embeddings are learned. Character and word encoders have 50 and 100 hidden units, respectively, while the decoder LSTM has a hidden size of 50. Dropout is used after every LSTM, as well as for word embedding input. We use Adam as an optimizer. Our model is built using MXNet. Hyperparameters are tuned using Bayesian Optimization \cite{snoek2012practical}.

\subsection{RE Model Settings}
Our final network had two encoder layers, with 8 attention heads in each multi-head attention sublayer and 256 filters for convolution layers in position-wise feedforward sublayer. We used dropout with probability 0.3 after the embedding layer, head/tail MLPs and the output of each encoder sublayer. We also used a word dropout with probability 0.15 before the embedding layer.

\section{Results}
\label{sec:results}
\subsection{NER and Trait Detection Results}
We report the results for NER and negation detection for both the i2b2 and DCN datasets in Table \ref{tab:results_ner}. We observe that our purposed conditional softmax decoder approach outperforms the best model \cite{chal2016ner} on the i2b2 challenge.

We compare our models for negation detection against NegEx \cite{negex} and ABoW \cite{shivade2015extending}, which has the best results for the negation detection task on i2b2 dataset. Conditional softmax decoder model outperforms both NegEx and ABoW (Table \ref{tab:results_ner}). Low performance of NegEx and ABoW is mainly attributed to the fact that they use ontology lookup to index findings and negation regular expression search within a fixed scope. A similar trend was observed in the medication condition dataset (Table \ref{tab:results_ner}). The important thing to note is the low F1 score for NegEx. This can primarily be attributed to abbreviations and misspellings in clinical notes which can not be handled well by rule-based systems.

\begin{table}[ht!]
    \caption{Test set performance with multi-task i2b2 and DCN datasets}
    \label{tab:results_ner}
    \centering
    \begin{tabular}{|l|l|l|l|l|}
        \hline
        \textbf{Data} & \textbf{Model} & \textbf{Precision} & \textbf{Recall} & \textbf{$\text{F}_1$}\\ 
       
        \hline
        \multicolumn{5}{c}{Named Entity}\\
        \hline
        \multirow{2}{*}{i2b2} & LSTM:CRF \cite{chal2016ner} & 0.844 & 0.834 & 0.839  \\
        & \textbf{Conditional Decoder} & 0.854 & 0.858 & \textbf{0.855}\\
        \hline
        \multirow{2}{*}{DCN} & LSTM:CRF \cite{chal2016ner} & 0.82 & 0.84 & 0.83\\
        & \textbf{Conditional Decoder} & 0.878 & 0.872 & \textbf{0.874}\\
        \hline
        \multicolumn{5}{c}{Negation}\\
        \hline
        \multirow{2}{*}{i2b2} & Negex \cite{negex}  &  0.896 & 0.799 & 0.845 \\
        & ABoW Kernel \cite{shivade2015extending} & 0.899 & 0.900 & 0.900 \\
        & \textbf{Conditional Decoder} & 0.919 &   0.891 & \textbf{0.905}\\
        \hline
        \multirow{2}{*}{DCN} & Negex \cite{negex} & 0.403 & 0.932 & 0.563 \\
        & \textbf{Conditional Decoder} & 0.928 &  0.874 & \textbf{0.899}\\
        \hline
    \end{tabular}
\end{table}

We also evaluated the conditional softmax decoder in low resource settings, where we used a sample of our training data. We observed that conditional decoder is more robust in low resource settings than other approaches as we reported in \cite{Bhatia2019}.

\subsection{RE Results}
To show the benefits of using second-order relations we compared our model’s performance to BRAN. The two models are different in the weighted addition of second-order relation scores. We tune over this weight parameter on the dev set and observed an improvement in MacroF1 score from 0.712 to 0.734 over DCN data and from 0.395 to 0.407 over i2b2 data. For further comparison a recently published model called Hybrid Deep Learning Approach (HDLA) \cite{chikka2018hybrid} reported a macroF1 score of 0.388 on the same i2b2 dataset. It should be mentioned that HDLA used syntactic parsers for feature extraction but we do not use any such external tools. 

Table \ref{tab:results_rel} summarizes the performance of our relationship model (+SOR) using second-order relations compared to BRAN and HDLA. We refer the readers to \cite{gaurav2019} for more detailed analysis of our relationship extraction model.

\begin{table}[ht!]
    \caption{Test set performance of relation extraction on i2b2 and DCN datasets}
    \label{tab:results_rel}
    \begin{center}
    \begin{tabular}{|l|l|l|l|l|}
    \hline
       \textbf{Data} & \textbf{Model} & \textbf{Precision} & \textbf{Recall} & \textbf{$\text{F}_1$}\\ 
       \cline{1-5}
        \multirow{2}{*}{i2b2} & HDLA \cite{chikka2018hybrid} & 0.378 & 0.422 & 0.388 \\
        & BRAN \cite{verga-etal-2018-simultaneously} & 0.396 & 0.403 & 0.395 \\
        & +SOR & 0.424 & 0.419 & \textbf{0.407} \\
       \hline
       \multirow{2}{*}{DCN} & BRAN \cite{verga-etal-2018-simultaneously} & 0.614 & 0.85 & 0.712 \\
        & +SOR & 0.643 & 0.879 & \textbf{0.734} \\
     \hline
    \end{tabular}
   \end{center}
\end{table}

\begin{figure*}[ht!]
\centering
    \centerline{\includegraphics[scale=0.45]{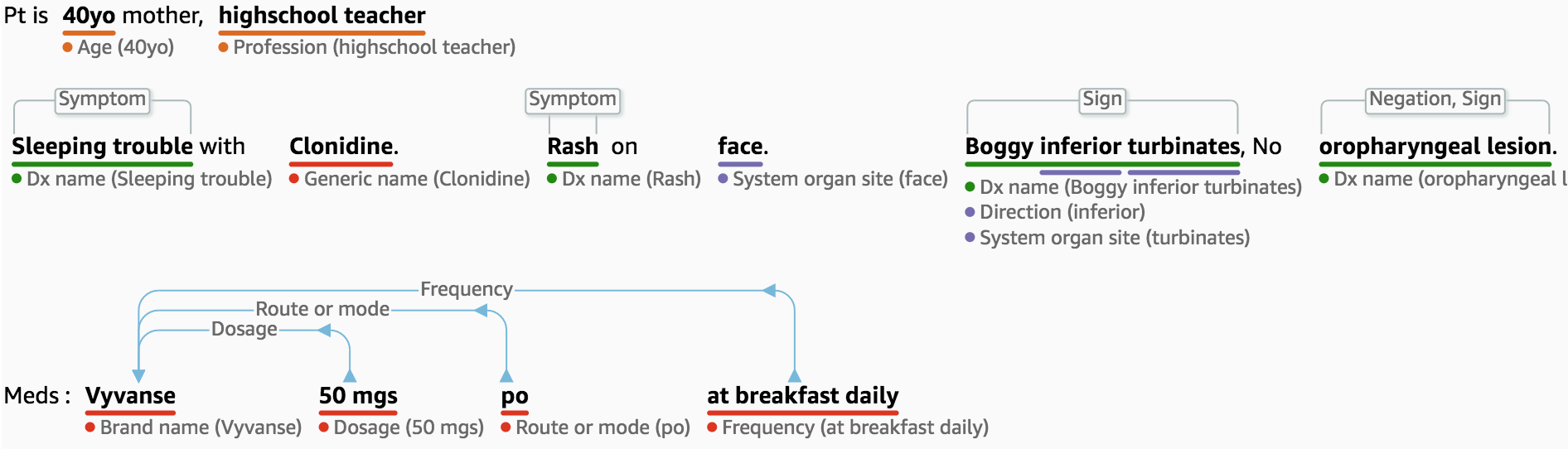}}
    \caption{Rendering of entities, traits and relations by Comprehend Medical UI}
    \label{fig:ui}
\end{figure*}

\section{Implementation}
\label{sec:imp}
Comprehend Medical APIs run in Amazon's proven, high-availability data centers, with service stack replication configured across three facilities in each AWS region to provide fault tolerance in the event of a server failure or Availability Zone outage. Additionally, Comprehend Medical ensures that system artifacts are encrypted in transit and user data is pass through and will not be stored in any part of the system.

\begin{figure*}[ht!]
    \centering
    \centerline{\includegraphics[scale=0.5]{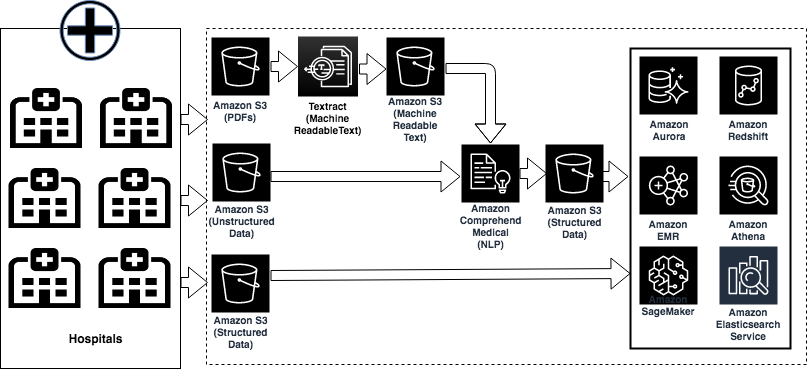}}
    \caption{Integrating Comprehend Medical into data processing pipeline}
    \label{fig:use_case}
\end{figure*}

Comprehend Medical is available through a Graphical User interface (GUI) within the AWS console and can be accessed using the Java and Python SDK. Comprehend Medical offers two APIs: 1) the NERe API which returns all the extracted named entities, their traits and the relationships between them, 2) the PHId API which returns just the protected health information contained in the text. Developers can easily integrate Comprehend Medical into their data processing pipelines as shown in Fig. \ref{fig:use_case}.

The only input needed by Comprehend Medical is the text to be analyze. No configuration, customization or other parameters needed, making Comprehend Medical easy to use by anyone who has access to AWS. Comprehend Medical outputs the results in JavaScript Object Notation (JSON), which contains named entities, begin offset, end offset, traits, confidence scores and the relationships between the entities. Using the GUI (Fig. \ref{fig:ui}) users can quickly visualize their results.

\section{Entities, Traits and Relationships}
\label{sec:entities}
\subsection{Entities}
Named entity mentions found in narrative notes are tagged with entity types listed in Table \ref{tab:list_of_entities}. The entities are divided into five categories: Anatomy, Medical Condition, Medication, PHI and TTP. Comprehend Medical is HIPAA eligible and therefore it supports HIPAA identifiers. Some of those identifiers are grouped under one identifier. For instance, Contact Point covers phone and fax numbers, and ID covers social security number, medical record number, account number, certificate or license number and vehicle or device number. An example input text is shown in Fig. \ref{fig:ui}.

\begin{table}[ht!]
  \caption{Entities extracted by Comprehend Medical}
  \label{tab:list_of_entities}
  \begin{center}
  \begin{tabular}{|l|l|}
  \hline
  \textbf{Category}&\textbf{Entity} \\
  \cline{1-2} 
    \multirow[t]{2}{*}{Anatomy} & Direction\\
                                & System Organ Site \\
    \hline
    \multirow[t]{2}{*}{Medical Condition}    & Dx Name \\
                                & Acuity \\
    \hline
    \multirow[t]{9}{*}{Medication}           & Brand Name \\
                                & Generic Name \\
                                & Dosage \\
                                & Duration \\
                                & Frequency \\
                                & Form \\
                                & Route or Mode \\
                                & Strength \\
                                & Rate \\
    \hline
    \multirow[t]{9}{*}{PHI}        
                                & Age \\
                                & Date \\
                                & Name \\
                                & Contact Point \\
                                & Email \\
                                & URL \\
                                & Identifier \\
                                & Address \\
                                & Profession \\
    \hline
    \multirow[t]{5}{*}{TTP}
                                & Test Name \\
                                & Test Value \\
                                & Test Unit \\
                                & Procedure Name \\
                                & Treatment Name \\
    \hline
\end{tabular}
\end{center}
\end{table}

\subsection{Traits}
Comprehend Medical covers four traits, listed in Table \ref{tab:list_of_traits}. Negation asserts the presence or absence of a Dx Name and whether or not the individual is taking the medication. Dx Name has three additional traits: Diagnosis, Sign and Symptom. {\bf Diagnosis} identifies an illness or a disease. {\bf Sign} is an objective evidence of disease and it is a phenomenon that is detected by a physician or a nurse. {\bf Symptom} is a subjective evidence of disease and it is phenomenon that is observed by the individual affected by the disease. An example of traits is shown in Fig. \ref{fig:ui}.

\begin{table}[ht!]
  \caption{Traits extracted by Comprehend Medical}
  \label{tab:list_of_traits}
  \begin{center}
  \begin{tabular}{|l|l|}
  \hline
  \textbf{Trait}&\textbf{Entity} \\
  \cline{1-2} 
    Negation & Brand/Generic Name, Dx Name \\
    Diagnosis & Dx Name \\
    Sign & Dx Name \\
    Symptom & Dx Name \\
    \hline
\end{tabular}
\end{center}
\end{table}

\subsection{Relationships}
A relationship is defined between a pair of entities in the Medication and TTP categories (Table \ref{tab:list_of_relations}). One of the entities in a relationship is the head while the other is the tail entity. In Medication, Generic and Brand Name are the head entity, which can have relationships to tail entities such as Strength and Dosage. An example of relations is shown in Fig. \ref{fig:ui}.

\begin{table}[ht!]
  \caption{Relationships extracted by Comprehend Medical}
  \label{tab:list_of_relations}
  \begin{center}
  \begin{tabular}{|l|l|}
  \hline
  \textbf{Head Entity}&\textbf{Tail Entity} \\
  \cline{1-2} 
    \multirow[t]{6}{*}{Brand/Generic Name}  & Dosage \\
                                    & Duration \\
                                    & Frequency \\
                                    & Form \\
                                    & Route or Mode \\
                                    & Strength \\
    \hline
    \multirow[t]{2}{*}{Test Name}   & Test Value \\
                                    & Test Unit \\
    \hline
\end{tabular}
\end{center}
\end{table}

\section{Use Cases}
\label{sec:use_cases}
Comprehend Medical reduces the cost, time and effort of processing large amounts of unstructured medical text with high accuracy, making it possible to pursue use cases such as clinical trial management, clinical decision support and revenue cycle management.

\subsection{Clinical Trial Management}
It can take about 10-15 years for a treatment to be developed from discovery to registration with the Federal Drug Administration (FDA). During that time, research organization can spend six years on clinical trails. Despite the number of year it takes to design those clinical trails, 90\% of all clinical trails fail to enroll patients within the targeted time and are forced to extend the enrollment period, 75\% of trails fail to enroll the targeted number of patients and 27\% fail to enroll any subjects \cite{giffin2010transforming}.

Life sciences and clinical research organizations can speed up and optimize the process of recruiting patients into a clinical trial as extractions from unstructured text and medical records can expedite the matching process. For instance, indexing patients based on medication, medical condition and treatments can help with quickly identifying the right participants for a lifesaving clinical trial.

Fred Hutchinson Cancer Research Center (FHCRC) utilized Comprehend Medical in their clinical trail management. FHCRC was spending 1.5 hours to annotate a single patient note, about 2.5 hours on manual chart abstraction per patient and per day they can process charts for about three patients. By using Comprehend Medical, FHCRC was able to annotate 9,642 patient notes per hour.

\subsection{Patient and Population Health Analytics}
Population health focuses on the discovery of factors and conditions for the a health of a population over time. It aims at identifying patterns of occurrence and knowledge discovery in order to develop polices and actions to improve health of a group or population \cite{Giannangelo2008}. 

Examples of population health analytics include patient stratification, readmission prediction and mortality measurement. Automatically unlocking important information from the narrative text is invaluable to organizations participating in value-based healthcare and population health. Structured medical records do not fully identify patients with medical history of diabetes, which results in an underestimation of disease prevalence \cite{Zheng2016}.  The inability to identify patient cohorts from structured data represents a problem for the development of population health and clinical management systems. It also negatively affects the accuracy of identifying high-risk and high-cost patients \cite{Bates2014}. Ref. \cite{Greenwald2017} identified three areas that may have an impact on readmission, but that are poorly documented in the EMR system, thus the need for NLP-based solutions to extract such information. Also, some symptoms and illness characteristics that are necessary to develop reliable predictors are missing in the coded billing data \cite{Rumshisky2016}. Ref. \cite{Jin2018} performed mortality prediction and reported a 2\% increase in the Area Under the Curve when using features from both structured data and concepts extracted from narrative notes and \cite{Poulin2014} found that the predictive power of suicide risk factors found in EMR systems become asymptotic, leading them to incorporate analysis on clinical notes to predict risk of suicide. 

As seen from the examples above, NLP-based approaches can assist in identifying concepts that are incorrectly codified or are missing in EMR system. Population health platforms can expand their risk analytics to leveraged unstructured clinical data for prediction of high risk patients and epidemiologic studies on outbreaks of diseases.

\subsection{Revenue Cycle Management}
In healthcare, Revenue Cycle Management (RCM) is the process of collecting revenue and tracking claims from healthcare providers including hospitals, outpatient clinics, nursing homes, dentist clinics and physician groups \cite{mindel2015contextualist}.

RCM process has been inefficient as most healthcare systems use rule-based approaches and manual audits of documents for billing and coding purposes \cite{Schouten2013}. Rule-based systems are time consuming, expensive to maintain, require attention and frequent human intervention. Due to these ineffective processes, data coded at point care, which is the source for claims data, can contain errors and inconsistencies. 

Coding is the process of encoding the details of patient encounters into standardized terminology \cite{Giannangelo2008}. A study by \cite{Ballaro2000} shows that 48 errors found in 38 of the 106 finished consultant episodes in urology and 71\% of these errors are caused by inaccurate coding. Ref. \cite{Lorence2003} measured the consistency of coded data and found that some of these errors were significant enough to change the diagnostic related group.

RCM companies can use Comprehend Medical to enhance existing workflows around computer assisted coding, and validate submitted codes by providers. In addition, claim audits, which often requires finding text evidence for submitted claims and is done manually, could be done more accurately and faster. 

\subsection{Pharmacovigilance}
The aim of pharmacovigilance is to monitor, detect and prevent adverse drug events (ADE) of medical drugs. Early system used for pharmacovigilance is the spontaneous reporting system (SRS), which provided safety information on drugs \cite{Wang2009}. However, SRS databases are incomplete, inaccurate and contain biased reporting \cite{Wang2009,Henriksson2015}. A newer generation of databases was created that contains clinical information for large patient population, such as the Intensive Medicines Monitoring Program (IMMP) and the General Practice Research Database (GPRD). Such databases included data from structured fields and forms, but very small amount of details are stored in the structured fields. Researchers then started to look into EHR data for pharmacovigilance. However, most valuable information in patient records are contained in the unstructured text.

Using NLP to extract information from narrative text have shown improvement in ADE detection and pharmacovigilance \cite{Luo2017}. Ref. \cite{Henriksson2015, Shang2014} also reported that ADEs are underreported in EHR systems and they used NLP techniques to enhance ADE detection.

\section{Conclusion}
\label{sec:conclusion}
Studies have shown that narrative notes are more expressive, more engaging and captures patient's story more accurately compared to the structured EHR data. They also contain more naturalistic prose, more reliable in identifying patients with a given disease and more understandable to healthcare providers reviewing those notes, which urges the need for a more accurate, intuitive and easy to use NLP system. In this paper we presented Comprehend Medical, a HIPAA eligible Amazon Web Service for medical language entity recognition and relationship extraction. Comprehend Medical supports several entity types divided into five different categories (Anatomy, Medical Condition, Medication, Protected Health Information, Treatment, Test and Procedure) and four traits (Negation, Diagnosis, Sign, Symptom). Comprehend Medical uses state-of-the-art deep learning models and provides two APIs, the NERe and PHId API. Comprehend Medical also comes with four different interfaces (CLI, Java SDK, Python SDK and GUI) and contrary to many other existing clinical NLP systems, it does not require dependencies, configuration or pipelined components customization.

\bibliography{ref}
\bibliographystyle{IEEEtran}

\end{document}